\documentclass[conference]{IEEEtran}
\IEEEoverridecommandlockouts
\usepackage{cite}
\usepackage{orcidlink}

\usepackage{amsmath,amssymb,amsfonts}
\usepackage{algorithmic}
\usepackage{graphicx}
\usepackage{textcomp}
\usepackage{xcolor}
\usepackage{comment}
\usepackage{hyperref}
\usepackage[acronym]{glossaries}
\usepackage{xspace}
\newcommand{\DSMV}{\textsc{DSM-V}\xspace}
\newcommand{\ICDten}{\textsc{ICD-10}\xspace}
\newcommand{\DualKB}{\textsc{Dual-KB}\xspace}
\newcommand{\NoKB}{\textsc{No-KB}\xspace}
\usepackage{booktabs,multirow,makecell}
\def\BibTeX{{\rm B\kern-.05em{\sc i\kern-.025em b}\kern-.08em
    T\kern-.1667em\lower.7ex\hbox{E}\kern-.125emX}}

\newacronym{ai}{AI}{Artificial Intelligence}
\newacronym{ml}{ML}{Machine Learning}
\newacronym{llm}{LLM}{Large Language Model}
\newacronym{rag}{RAG}{Retrieval-Augmented Generation}
\newacronym{kb}{KB}{Knowledge base}

\newacronym{dsm5}{DSM-V}{Diagnostic and Statistical Manual of Mental Disorders, Fifth Edition}
\newacronym{icd10}{ICD-10}{International Classification of Diseases, Tenth Revision}
\newacronym{icd11}{ICD-11}{International Classification of Diseases, Eleventh Revision}

\newacronym{gan}{GAN}{Generative Adversarial Network}
\newacronym{vae}{VAE}{Variational Autoencoder}

\newacronym{dp}{DP}{Differential Privacy}
\newacronym{jsd}{JSD}{Jensen–Shannon Divergence}
\newacronym{cs}{CS}{Chi-Squared Test}
\newacronym{qi}{QI}{Quasi-Identifiers}

\newacronym{ehr}{EHR}{Electronic Health Record}
\newacronym{tre}{TRE}{Trusted Research Environment}
\newacronym{who}{WHO}{World Health Organization}
\begin{document}

\title{Knowledge-Guided Retrieval-Augmented Generation for Zero-Shot Psychiatric Data: Privacy Preserving Synthetic Data Generation}

\author{
    \IEEEauthorblockN{
        Adam Jakobsen\orcidlink{0009-0005-0264-3119}\IEEEauthorrefmark{1}\IEEEauthorrefmark{2},  
        Sushant Gautam\orcidlink{0000-0001-9232-2661}\IEEEauthorrefmark{1}\IEEEauthorrefmark{2},  
        Hugo Lewi Hammer\orcidlink{0000-0001-9429-7148}\IEEEauthorrefmark{1}\IEEEauthorrefmark{2}
        Susanne Olofsdotter\orcidlink{0000-0002-8114-8386}\IEEEauthorrefmark{3}
        Miriam S Johanson\orcidlink{0000-0001-7486-2859}\IEEEauthorrefmark{2}\\
        Pål Halvorsen\orcidlink{0000-0003-2073-7029}\IEEEauthorrefmark{1}\IEEEauthorrefmark{2},  
        Vajira Thambawita\orcidlink{0000-0001-6026-0929}\IEEEauthorrefmark{1}, 
    }
    \IEEEauthorblockA{
        \IEEEauthorrefmark{1}SimulaMet, Norway \quad
        \IEEEauthorrefmark{2}Oslo Metropolitan University, Norway \quad
        \IEEEauthorrefmark{3}Uppsala University, Sweden}
    }

\maketitle

\begin{abstract}
AI systems in healthcare research have shown potential to increase patient throughput and assist clinicians, yet progress is constrained by limited access to real patient data. To address this issue, we present a zero-shot, knowledge-guided framework for psychiatric tabular data in which large language models (LLMs) are steered via Retrieval-Augmented Generation using the Diagnostic and Statistical Manual of Mental Disorders (DSM-5) and the International Classification of Diseases (ICD-10). We conducted experiments using different combinations of knowledge bases to generate privacy-preserving synthetic data. The resulting models were benchmarked against two state-of-the-art deep learning models for synthetic tabular data generation, namely CTGAN and TVAE, both of which rely on real data and therefore entail potential privacy risks. Evaluation was performed on six anxiety-related disorders: specific phobia, social anxiety disorder, agoraphobia, generalized anxiety disorder, separation anxiety disorder, and panic disorder. 
CTGAN typically achieves the best marginals and multivariate structure, while the knowledge-augmented LLM is competitive on pairwise structure and attains the lowest pairwise error in separation anxiety and social anxiety. An ablation study shows that clinical retrieval reliably improves univariate and pairwise fidelity over a no-retrieval LLM. Privacy analyses indicate that the real data-free LLM yields modest overlaps and a low average linkage risk comparable to CTGAN, whereas TVAE exhibits extensive duplication despite a low $k$-map score. Overall, grounding an LLM in clinical knowledge enables high-quality, privacy-preserving synthetic psychiatric data when real datasets are unavailable or cannot be shared.
\end{abstract}


\section{Introduction}
Mental health disorders represent a significant and growing global health challenge, affecting a substantial portion of the population \cite{Moitra2023}. According to the \gls{who}, one in every eight people worldwide lives with a mental disorder \cite{who2022mental,coombs2021barriers}. \gls{ai} and \gls{ml} have emerged as powerful paradigms with the potential to accelerate the 
diagnostic process, enhance diagnostic precision, and improve access to mental healthcare \cite{Rony2025,graham2019artificial}. However, progress in this domain is severely constrained by limited access to large-scale, high-quality psychiatric data. The sensitive nature of this information requires strict adherence to privacy regulations, creating a bottleneck for research and model development~\cite{Du2023}.

To circumvent these challenges, deep learning synthetic data generators have emerged as a promising solution to data scarcity in healthcare \cite{Choi2021,Giuffre2023}. While powerful, these methods are fundamentally reliant on access to large volumes of real patient data for training. This dependency introduces a utility-privacy paradox. To protect patient confidentiality, techniques like \gls{dp}, where noise is injected during the learning process, are often applied during training of the generative model \cite{dwork2014algorithmic}. However, this injected noise can degrade the statistical integrity of the learned data distribution, consequently diminishing the utility of the synthetic data for downstream tasks
\cite{van2023generating,zhang2021privsyn}. This inherent trade-off between 
privacy and utility limits the effectiveness of data-dependent generation methods, as stronger privacy guarantees often lead to less useful synthetic data.

In this work, we introduce a fundamentally different paradigm for psychiatric data synthesis that completely circumvents the data-dependency problem and its associated fidelity-utility-privacy paradox. To generate privacy-preserving synthetic data, we propose a zero-shot, knowledge-guided framework that leverages a \gls{llm} to ``simulate" a patient completing a structured clinical assessment based on the \gls{dsm5}~\cite{dsm}. By employing \gls{rag} \cite{lewis2020retrieval}, our model grounds its responses in a knowledge base of clinical criteria, enabling it to generate clinically plausible and coherent assessment data without being trained on or exposed to real patient records. This zero-shot approach breaks the reliance on sensitive data, allowing for the creation of high-fidelity, privacy-preserving synthetic datasets from first principles. We evaluate our generated datasets in terms of both fidelity and privacy, demonstrating a viable path toward accelerating \gls{ai} research in mental health without compromising patient privacy. Source code and datasets are available at \url{https://github.com/Adamjakobsen/Zero-Shot-Psychiatry-with-RAG}.

\section{Related Work}
Our work is situated at the intersection of three rapidly advancing domains: synthetic data generation in healthcare, privacy-preserving machine learning, and the application of \glspl{llm} for clinical simulation. This section reviews key developments in these areas and contextualizes the novelty of our knowledge-guided, zero-shot approach.

\subsection{Synthetic Data Generation for Health Data}
The use of synthetic data to overcome privacy and accessibility barriers in healthcare is a well-established field of research. \gls{gan}-based models, such as MedGAN \cite{choi2017generating}, have been successfully used to generate realistic synthetic \gls{ehr}. To better handle the mixed-type nature of tabular data common in EHRs, specialized architectures like the Conditional Tabular GAN (CTGAN) were developed \cite{ctgan}. CTGAN and its variants have become a popular baseline for generating tabular medical data, demonstrating strong performance in preserving the statistical properties of the original data across various medical applications \cite{das2023generating}. Subsequent work has extended \gls{gan} architectures to handle the temporal nature of longitudinal patient data, as seen in models like RCGAN \cite{esteban2017real} and DoppelGANger
\cite{lin2020using}. VAEs have also been employed for similar tasks, often noted 
for their more stable training dynamics, with models like MedVAE and TVAE demonstrating strong performance in generating plausible patient records \cite{biswal2021medvae,ctgan}. While these models excel at learning distributions from existing data, their fundamental limitation is their complete dependence on a large, representative real dataset for training. This reliance makes them vulnerable to privacy risks and perpetuates data scarcity issues when no initial dataset is available. Our work diverges from this entire paradigm by eliminating the need for a real training dataset altogether.

\subsection{Privacy Preservation in Synthetic Health Data}
To mitigate the privacy risks associated with training generative models on sensitive data, Differential Privacy (DP) has become the gold standard
\cite{dwork2014algorithmic}. The core idea of DP is to add carefully calibrated 
noise at some point in the model training pipeline—either to the input data, the model gradients (as in DP-SGD \cite{abadi2016deep}), or the output—to make it computationally infeasible to determine whether any single individual's data was used in the training.

Numerous studies have integrated DP with generative models for health data. For example, DP-GAN~\cite{xie2018differentially} applies noise to the gradients of the discriminator during GAN training to provide formal privacy guarantees. Similarly, PATE-GAN~\cite{jordon2018pate} uses the Private Aggregation of Teacher Ensembles (PATE) framework to achieve privacy. However, this formal privacy comes at a cost. The noise injection required for DP inherently creates a trade-off between privacy, fidelity and data utility~\cite{kaissis2021scoping,zhang2021privsyn}. High levels of privacy (i.e., more noise) often lead to a degradation in the statistical fidelity of the synthetic data, reducing its effectiveness for training downstream machine learning models. Our knowledge-guided approach sidesteps this fidelity-privacy paradox by generating data from a knowledge base rather than a private dataset, thus providing privacy by design without sacrificing utility.

\subsection{LLM for Clinical Data Synthesis and Simulation}
More recently, the capabilities of \glspl{llm} have been explored for clinical data synthesis. Models like GatorTron \cite{yang2022large} and Med-PaLM \cite{singhal2023large} have been fine-tuned on massive corpora of clinical text, enabling them to generate realistic clinical notes and patient summaries. These approaches, however, are still a form of data-dependent synthesis. They rely on models that have been pre-trained or fine-tuned on vast amounts of clinical data, which carries an implicit risk of memorization and patient information leakage \cite{carlini2021extracting}.

A closer parallel to our work is the use of \glspl{llm} as simulators. For instance,
Park et al. \cite{agnew2023generative} demonstrated that \glspl{llm} can simulate realistic human behavior. In the clinical domain, \glspl{llm} have been used to simulate doctor-patient conversations for training purposes~\cite{Haider2025}. Our work builds on this concept of simulation but applies it in a novel, knowledge-grounded manner. By using \gls{rag}~\cite{lewis2020retrieval} to dynamically pull in clinical knowledge from \gls{dsm5} and \gls{icd10}~\cite{icd}, we constrain 
the LLM's generation process, ensuring that the simulated patient responses are not just fluent but also clinically plausible and consistent with established diagnostic knowledge. This zero-shot, knowledge-guided simulation represents a new frontier for creating privacy-by-design synthetic psychiatric data.

\section{The approach: Knowledge-Guided Data Generation}


This work presents a novel approach for generating synthetic psychiatric tabular data using \glspl{llm} enhanced with \gls{rag}. Our approach combines the clinical knowledge retrieval capabilities of RAG with the natural language generation power of \glspl{llm} to simulate realistic patient self-assessments for psychiatric disorders. To evaluate our approach, we use an open-access clinical dataset from OSF (\url{https://osf.io/jz4ge/}\cite{VidalArenas2023DSM5}) consisting of \gls{dsm5} Severity measures from $564$ adult participants. The responses from the participants consist of discrete Likert scores $s_{pdi}\in\{0,1,2,3,4\}$, where 0=\textit{never} and 4=\textit{all of the time}. For this study, we treat each DSM-5 severity scale independently, producing six disorder-specific tables: Separation Anxiety, Specific Phobia, Social Anxiety, Panic, Agoraphobia, and Generalized Anxiety. Additionally, we restrict analyses to the baseline assessment; thus, we do not model comorbidity or longitudinal change. From a clinical perspective, the decision to model disorders independently is a necessary first step, but it is a significant simplification of psychiatric practice, where comorbidity is the rule rather than the exception. Missing responses are imputed using Multiple Imputation by Chained Equations (MICE) with a Bayesian ridge estimator\cite{Azur2011Mar,Tipping2001Sep}. The resulting real and synthetic datasets are available at \url{https://github.com/Adamjakobsen/Zero-Shot-Psychiatry-with-RAG}.

\subsection{System Architecture}
\begin{figure*}[!t]
    \centering
    \includegraphics[width=\linewidth]{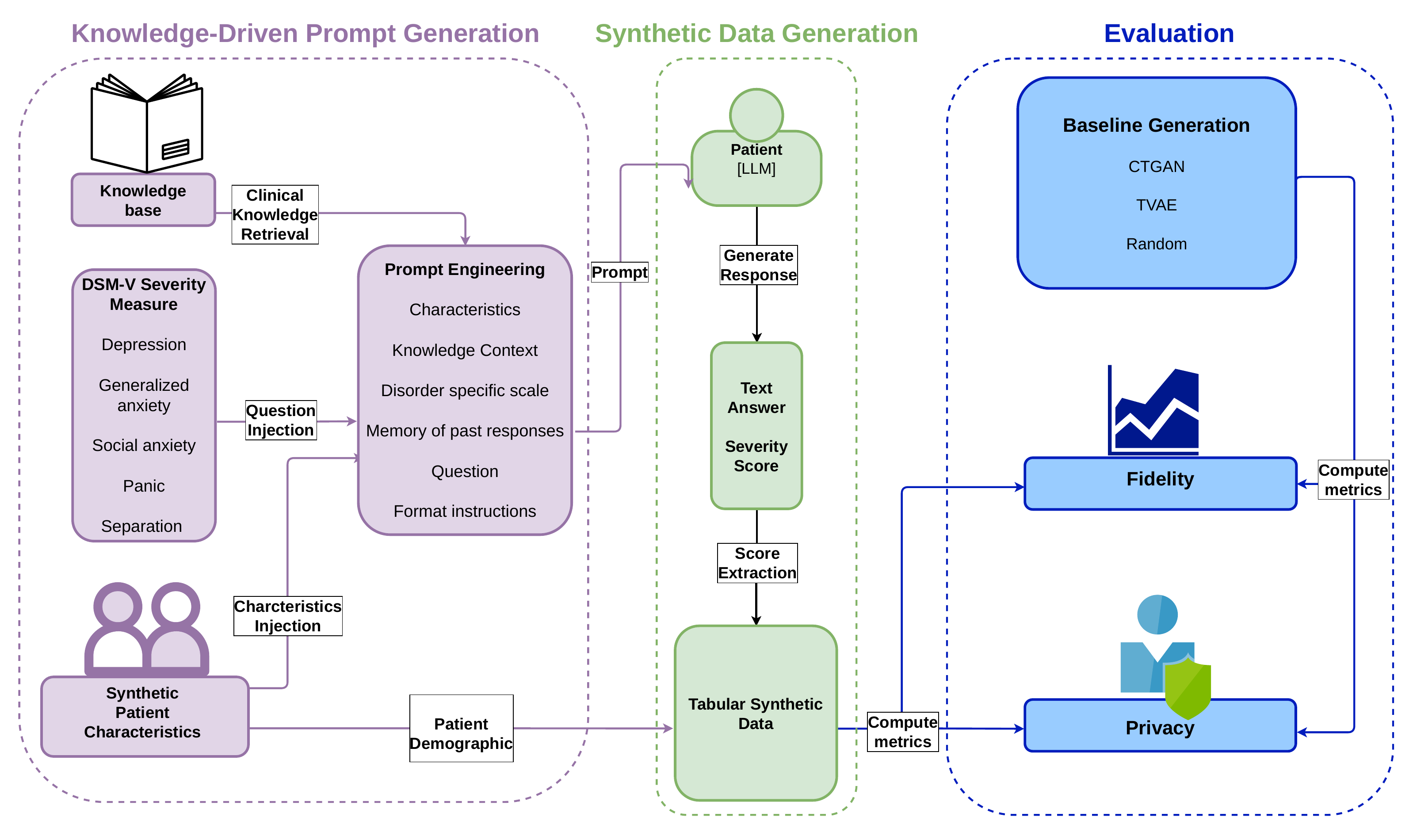}
    \caption{Pipeline of the proposed framework. Synthetic patients are generated by conditioning a Large Language Model (LLM) on sampled personas and clinical knowledge retrieved via Retrieval-Augmented Generation (RAG). Outputs are tabular datasets evaluated for fidelity and privacy against baselines. Abbreviations: LLM, Large Language Model; RAG, Retrieval-Augmented Generation.}
    \label{fig:framework}
\end{figure*}

We synthesize item–level psychiatric tabular data using Mistral $7$B~\cite{jiang2023mistral7b}, a 7-billion-parameter \glspl{llm}, grounded by \gls{rag}. No real patient data are used at any stage. Figure~\ref{fig:framework} summarizes the pipeline.
\textit{Knowledge base and retrieval:}
We build a vector store using the Faiss library \cite{faiss} from clinical references and evaluate four modes:
\NoKB\ (no retrieval), \DSMV, \ICDten, and \DualKB\ (both). For a given disorder $d$, the retriever returns the top-$k$ clinically relevant snippets (definitions, symptom criteria, and item wording) that are injected into the prompt as \emph{Knowledge Context}.
\textit{Patient characteristics sampling:}
For each synthetic patient $p$, we sample demographics and communication traits:
sex $\in\{\text{female},\text{male}\}$, age $\in \mathbb{N}$, and a latent response profile (severity prior, response style, symptom awareness, communication style, and a consistency level). These variables condition the LLM to produce coherent, patient-like responses over the questionnaire.

For each disorder $d$ with item set $\{q_i\}_{i=1}^{N_d}$, we run a conversational loop. The prompt contains:
\begin{enumerate}
    \item The sampled patient characteristics
    \item The retrieved Knowledge Context (if any)
    \item The current item $q_i$ from the DSM-5 self–report severity measure
    \item Memory of previous answers to enforce intra-patient consistency
    \item Formatting instructions 
\end{enumerate}

The LLM returns a free-text answer and a discrete Likert score $s_{pdi}\in\{0,1,2,3,4\}$ where 0=\textit{never} and 4=\textit{all of the time}. We deterministically extract $s_{pdi}$ with a schema-constrained parser.

\textit{Tabular construction:} Each conversation yields one row per synthetic patient and one dataset per disorder, such that
\begin{equation*}
\text{row}_{p,d}=\big[\texttt{sex},\ \texttt{age},\ \{d_\texttt{it}1,\dots,d_\texttt{it}{N_d}\}\big],
\quad 
\end{equation*}

where $ d_\texttt{it}i \equiv s_{pdi}$.


To evaluate the contribution of clinical knowledge retrieval, we implement a systematic ablation study by varying the knowledge base configuration. The four experimental conditions (\DualKB, \DSMV, \ICDten,
\NoKB) allow us to assess: (1) the overall impact of clinical knowledge 
on synthetic data quality, (2) the relative contributions of different diagnostic frameworks, and (3) the complementary effects of combining multiple knowledge sources.


\subsection{Fidelity Evaluation}
\label{sec:fidelity_eval}

All variables are categorical. For each disorder, let $D$ denote the real dataset (with $n$ rows and $p$ columns) and $\widehat{D}$ the synthesized dataset of identical row count. We quantify similarity by assessing statistical properties at three complementary levels as follows:

\begin{enumerate}
    
\item Univariate marginals:
To measure the similarity between the univariate distributions of the real and synthetic data, we employ the Jensen-Shannon Divergence (JSD) \cite{jensen1984shannon}. For each column $j\in\{1,\dots,p\}$ with empirical probability mass functions (PMFs) $\pi^{\text{real}}_j$ (from $D$) and $\pi^{\text{syn}}_j$ (from $\widehat{D}$),
\begin{equation*}
\begin{split}
\mathrm{JSD}(\pi^{\text{real}}_j,\pi^{\text{syn}}_j)
&=\tfrac{1}{2}\, \mathrm{KL}\!\Bigl(\pi^{\text{real}}_j \,\Big\|\, \tfrac{\pi^{\text{real}}_j+\pi^{\text{syn}}_j}{2}\Bigr)\\
&+\tfrac{1}{2}\, \mathrm{KL}\!\Bigl(\pi^{\text{syn}}_j \,\Big\|\, \tfrac{\pi^{\text{real}}_j+\pi^{\text{syn}}_j}{2}\Bigr),
\end{split}
\end{equation*}
and we summarize by the mean across columns, $\frac{1}{p}\sum_{j=1}^p \mathrm{JSD}(\pi^{\text{real}}_j,\pi^{\text{syn}}_j)$. Lower values indicate closer univariate agreement.

\item Bivariate associations:
For each unordered pair $(j,k)$, we form the real and synthetic contingency tables and compute bias-corrected Cramér’s~$V$ for $D$ and $\widehat{D}$; denote them $V^{\text{real}}_{jk}$ and $V^{\text{syn}}_{jk}$\cite{bergsma2013}. The mean absolute Cramér’s~$V$ error is then defined as:
\begin{equation*}
\mathrm{MAE_V} \;=\; 1 \;-\; \frac{2}{p(p-1)} 
\sum_{j<k} \Bigl|\,V^{\text{real}}_{jk} - V^{\text{syn}}_{jk}\Bigr|.
\end{equation*}

\item Multivariate structure:
Let $P$ and $Q$ denote the empirical multivariate distributions of $D$ and $\widehat{D}$, respectively. We compute the energy distance squared\cite{Szekely2013Aug}, denoted by $\mathrm{ED}^2(P,Q)$, using Hamming distance $\|u-v\|_H=\sum_{j=1}^p \mathbb{1}\{u_j\neq v_j\}$ on categorical vectors,
\begin{equation*}
\begin{split}
\mathrm{ED}^2(P,Q) \;=&\; 2\,\mathbb{E}\|X-Y\|_H 
\;-\; \mathbb{E}\|X-X'\|_H\\ 
&\;-\; \mathbb{E}\|Y-Y'\|_H,
\end{split}
\end{equation*}
with $X,X'\!\sim P$ and $Y,Y'\!\sim Q$, estimated by the standard V/U-statistic. Lower values indicate closer multivariate alignment.
\end{enumerate}

\subsection{Privacy Evaluation}
\label{sec:privacy_eval}

Exact Overlap (EO) is defined as the
fraction of synthetic rows that exactly coincide with at least one real row:
\begin{equation*}
\mathrm{EO}(\widehat{D},D)
= \frac{1}{|\widehat{D}|}\,\Big|\big\{\hat{\mathbf{x}}\in\widehat{D}:\;\hat{\mathbf{x}}\in D\big\}\Big|.
\end{equation*}
For each $\hat{\mathbf{x}}\in\widehat{D}$, the normalized nearest-neighbour distance is given as:
\begin{equation*}
d_{\mathrm{NN}}(\hat{\mathbf{x}},D)
=\frac{1}{p}\,\min_{\mathbf{x}\in D}\sum_{j=1}^{p}\mathbb{1}\{x_j\neq \hat{x}_j\}.
\end{equation*}
We report the 5th percentile $q_{0.05}\{d_{\mathrm{NN}}\}$ across $\widehat{D}$, where larger values indicate that even the closest synthetic records remain several categorical columns away from real records.

Let the \gls{qi} be $Q=\{\texttt{sex},\texttt{age}\}$. For a synthetic record with \gls{qi} pattern $q$, let $k(q)$ be the size of the real-data equivalence class $\{\mathbf{x}\in D:\mathbf{x}_Q=q\}$. The known-QI linkage risk for that record is $\mathbb{1}\{k(q)\!>\!0\}/k(q)$\cite{sweeney2002k}. We report the average over $\widehat{D}$:
\begin{equation*}
\mathrm{Risk}_{\text{k-map}}(\widehat{D},D)
=\frac{1}{|\widehat{D}|}\sum_{\hat{\mathbf{x}}\in\widehat{D}}
\frac{\mathbb{1}\{k(\hat{\mathbf{x}}_Q)>0\}}{k(\hat{\mathbf{x}}_Q)}.
\end{equation*}

\noindent

\subsection{Baselines for Synthetic Data Generation}
\label{sec:baselines}

We benchmark our zero-shot \gls{llm}+\gls{rag} generator against three baselines: CTGAN, TVAE, and a randomly sampled baseline. 

CTGAN and TVAE models are tuned with privacy-aware selection and trained using the SDV library~\cite{sdv}.
For each disorder, we create a fixed $70\%/30\%$ split of $D$ into $D_{\text{train}}$ and $D_{\text{tune}}$ (stratified on \texttt{sex} when available). All variables are treated as categorical. We conduct a small grid search over training epochs ($\in\{50,100,150,200,300\}$) and random seeds ($\in\{11,23,37\}$).
Each candidate model is trained on $D_{\text{train}}$ and used to sample $|\hat D|=|D_{\text{tune}}|$ rows for scoring. We compute one fidelity proxy on $D_{\text{tune}}$ (mean univariate $\mathrm{JSD}$) and three privacy proxies against $D_{\text{train}}$: Exact Overlap (EO), the near-match share $\mathrm{Share}(d_{\text{ham}}\le 1)$, and the nearest-neighbour $q_{0.05}$ of the \emph{unnormalized Hamming distance} $d_{\text{ham}}$ (the number of differing columns). We select the lowest-$\mathrm{JSD}$ candidate among those satisfying:
\begin{equation*}
\mathrm{EO}\le 0.01,\quad \mathrm{Share}(d_{\text{ham}}\le 1)\le 0.10,\quad q_{0.05}\{d_{\text{ham}}\}\ge 1.
\end{equation*}

If any candidate satisfied all three privacy gates, we choose the one with minimum $\mathrm{JSD}$. Otherwise, we fallback to: (i) minimize EO, then (ii) minimize near-match share, then (iii) maximize $q_{0.05}\{d_{\mathrm{ham}}\}$, then (iv) minimize $\mathrm{JSD}$. The selected configuration is then used to retrain the model on the full dataset with no additional privacy gates post-retraining.

Finally, we construct a randomly sampled dataset for each disorder to provide a fidelity and privacy floor driven solely by chance coincidences. Under realistic marginals, we draw \texttt{sex} and \texttt{age} from their empirical PMFs in $D$ and sample instrument items independently and uniformly within the range of choices given by the \DSMV questionnaires.

\section{Results and Analysis}
\label{sec:results}

\subsection{Fidelity comparison with baselines}
\label{sec:results:fidelity}

Table~\ref{tab:fidelity_results_complete} summarizes performance across the six anxiety disorders. As expected, models trained on real data set the upper bound on overall fidelity: \textsc{CTGAN} typically achieves the lowest $\mathrm{JSD}$ and $\mathrm{ED}^2$ across disorders, while \textsc{TVAE} attains the lowest pairwise–association error ($\mathrm{MAE}_V$) in Agoraphobia and Generalized Anxiety. The zero–shot LLM, when grounded in clinical knowledge, is competitive on pairwise structure and attains the best $\text{MAE}_V$ in two disorders—Separation Anxiety (\textit{\DualKB}: 0.148 [0.105, 0.193]) and Social Anxiety (\textit{\DualKB}: 0.163 [0.127, 0.199])—but generally trails \textsc{CTGAN} on $\mathrm{JSD}$/$\mathrm{ED}^2$ and \textsc{TVAE} on $\mathrm{MAE}_V$ in the remaining disorders. Within the LLM models, all knowledge–augmented variants outperform \textit{\NoKB} and random generation across metrics and disorders.

\begin{table*}[!t]
\centering
\footnotesize
\caption{Fidelity with 95\% bootstrap confidence intervals. Best results in \textbf{bold}.}

\label{tab:fidelity_results_complete}
\setlength{\tabcolsep}{6pt}
\renewcommand{\arraystretch}{1.1}
\begin{tabular}{llccc}
\toprule
\rotatebox{0}{\textbf{Disorder}} & \rotatebox{0}{\textbf{Method}} & \rotatebox{0}{\parbox[b]{1.35cm}{\centering \textbf{JSD} $\downarrow$}} & \rotatebox{0}{\parbox[b]{1.35cm}{\centering $\mathbf{MAE}_V$ $\downarrow$}} & \rotatebox{0}{\parbox[b]{1.35cm}{\centering \textbf{ED$^2$} $\downarrow$}} \\
\midrule
\multirow{7}{*}{\rotatebox[origin=c]{90}{\makecell{\strut Agoraphobia\strut}}} & CTGAN & \textbf{0.033 [0.025, 0.042]} & 0.249 [0.218, 0.279] & \textbf{0.033 [0.024, 0.043]} \\
  & TVAE & 0.070 [0.060, 0.081] & \textbf{0.159 [0.129, 0.192]} & 0.061 [0.049, 0.075] \\
  & \textit{\DSMV} & 0.082 [0.072, 0.094]\rlap{\textbf{*}}& 0.200 [0.132, 0.246]\rlap{\textbf{*}}& 0.097 [0.084, 0.112]\rlap{\textbf{*}}\\
  & \textit{\DualKB} & 0.083 [0.072, 0.095] & 0.207 [0.169, 0.246] & 0.097 [0.084, 0.113]\rlap{\textbf{*}}\\
  & \textit{\ICDten} & 0.094 [0.083, 0.106] & 0.212 [0.124, 0.269] & 0.101 [0.089, 0.116] \\
  & \textit{\NoKB} & 0.101 [0.087, 0.117] & 0.250 [0.139, 0.300] & 0.110 [0.094, 0.129] \\
  & Random & 0.242 [0.224, 0.261] & 0.225 [0.196, 0.253] & 0.287 [0.262, 0.312] \\
\midrule
\multirow{7}{*}{\rotatebox[origin=c]{90}{\makecell{\strut Generalized\\Anxiety\strut}}}& CTGAN & \textbf{0.017 [0.013, 0.022]} & 0.220 [0.177, 0.267] & \textbf{0.010 [0.007, 0.015]} \\
  & TVAE & 0.066 [0.057, 0.076] & \textbf{0.144 [0.118, 0.178]} & 0.055 [0.046, 0.066] \\
  & \textit{\DSMV} & 0.144 [0.127, 0.162] & 0.197 [0.146, 0.247] & 0.193 [0.167, 0.220] \\
  & \textit{\DualKB} & 0.146 [0.129, 0.164] & 0.155 [0.100, 0.212]\rlap{\textbf{*}}& 0.193 [0.168, 0.219] \\
  & \textit{\ICDten} & 0.138 [0.122, 0.156]\rlap{\textbf{*}}& 0.194 [0.124, 0.250] & 0.184 [0.159, 0.210]\rlap{\textbf{*}}\\
  & \textit{\NoKB} & 0.173 [0.154, 0.193] & 0.226 [0.176, 0.270] & 0.215 [0.190, 0.241] \\
  & Random & 0.174 [0.159, 0.192] & 0.214 [0.171, 0.260] & 0.186 [0.169, 0.204] \\
\midrule
\multirow{7}{*}{\rotatebox[origin=c]{90}{\makecell{\strut Panic\strut}}}
  & CTGAN & \textbf{0.059 [0.048, 0.072]} & 0.275 [0.234, 0.313] & \textbf{0.077 [0.060, 0.095]} \\
  & TVAE & 0.101 [0.090, 0.115] & \textbf{0.148 [0.115, 0.187]} & 0.106 [0.092, 0.122] \\
  & \textit{\DSMV} & 0.085 [0.076, 0.096] & 0.194 [0.147, 0.243] & 0.102 [0.091, 0.115] \\
  & \textit{\DualKB} & 0.082 [0.074, 0.092]\rlap{\textbf{*}}& 0.191 [0.137, 0.249] & 0.097 [0.087, 0.109]\rlap{\textbf{*}}\\
  & \textit{\ICDten} & 0.082 [0.074, 0.091]\rlap{\textbf{*}}& 0.153 [0.108, 0.196]\rlap{\textbf{*}}& 0.098 [0.088, 0.110] \\
  & \textit{\NoKB} & 0.095 [0.083, 0.109] & 0.203 [0.152, 0.253] & 0.105 [0.090, 0.123] \\
  & Random & 0.259 [0.237, 0.281] & 0.275 [0.234, 0.313] & 0.345 [0.313, 0.377] \\
\midrule
\multirow{7}{*}{\rotatebox[origin=c]{90}{\makecell{\strut Separation\\Anxiety\strut}}}  
  & CTGAN & \textbf{0.040 [0.031, 0.049]} & 0.205 [0.175, 0.233] & \textbf{0.049 [0.037, 0.062]} \\
  & TVAE & 0.107 [0.096, 0.119] & 0.177 [0.153, 0.203] & 0.109 [0.094, 0.125] \\
  & \textit{\DSMV} & 0.090 [0.080, 0.101] & 0.197 [0.123, 0.257] & 0.130 [0.113, 0.149] \\
  & \textit{\DualKB} & 0.096 [0.085, 0.108] & \textbf{0.148 [0.105, 0.193]}\rlap{\textbf{*}}& 0.141 [0.122, 0.161] \\
  & \textit{\ICDten} & 0.076 [0.068, 0.085]\rlap{\textbf{*}}& 0.232 [0.194, 0.270] & 0.093 [0.083, 0.106]\rlap{\textbf{*}}\\
  & \textit{\NoKB} & 0.095 [0.083, 0.107] & 0.239 [0.129, 0.276] & 0.128 [0.111, 0.148] \\
  & Random & 0.240 [0.221, 0.259] & 0.206 [0.176, 0.235] & 0.296 [0.271, 0.321] \\
\midrule
\multirow{7}{*}{\rotatebox{90}{Social Anxiety}} & CTGAN & \textbf{0.021 [0.016, 0.026]} & 0.233 [0.203, 0.263] & \textbf{0.013 [0.010, 0.018]} \\
  & TVAE & 0.146 [0.135, 0.157] & 0.207 [0.181, 0.237] & 0.146 [0.133, 0.160] \\
  & \textit{\DSMV} & 0.092 [0.080, 0.105] & 0.196 [0.155, 0.237] & 0.125 [0.106, 0.146] \\
  & \textit{\DualKB} & 0.088 [0.077, 0.100]\rlap{\textbf{*}}& \textbf{0.163 [0.127, 0.199]}\rlap{\textbf{*}}& 0.121 [0.103, 0.140] \\
  & \textit{\ICDten} & 0.093 [0.081, 0.106] & 0.185 [0.149, 0.220] & 0.119 [0.101, 0.138]\rlap{\textbf{*}}\\
  & \textit{\NoKB} & 0.114 [0.098, 0.130] & 0.230 [0.198, 0.263] & 0.150 [0.128, 0.173] \\
  & Random & 0.180 [0.163, 0.198] & 0.227 [0.198, 0.257] & 0.198 [0.179, 0.217] \\
\midrule
\multirow{7}{*}{\rotatebox{90}{Specific Phobia}} & CTGAN & \textbf{0.023 [0.017, 0.030]} & 0.183 [0.160, 0.207] & \textbf{0.023 [0.017, 0.030]} \\
  & TVAE & 0.187 [0.174, 0.201] & 0.215 [0.194, 0.237] & 0.190 [0.173, 0.209] \\
  & \textit{\DSMV} & 0.115 [0.101, 0.131] & 0.266 [0.238, 0.295] & 0.149 [0.129, 0.173] \\
  & \textit{\DualKB} & 0.113 [0.099, 0.129]\rlap{\textbf{*}}& 0.267 [0.238, 0.293] & 0.148 [0.127, 0.171]\rlap{\textbf{*}}\\
  & \textit{\ICDten} & 0.127 [0.112, 0.143] & 0.254 [0.228, 0.279]\rlap{\textbf{*}}& 0.165 [0.143, 0.190] \\
  & \textit{\NoKB} & 0.120 [0.105, 0.136] & 0.305 [0.279, 0.329] & 0.150 [0.130, 0.173] \\
  & Random & 0.131 [0.117, 0.146] & \textbf{0.180 [0.158, 0.205]} & 0.157 [0.141, 0.174] \\
\bottomrule
\end{tabular}
\end{table*}

\subsection{Ablation: effect of clinical retrieval on fidelity}
\label{sec:results:ablation}

We assess the impact of grounding the zero-shot LLM in clinical manuals by reporting paired bootstrap deltas against \NoKB for three fidelity metrics: $\mathrm{JSD}$, $\mathrm{MAE}_V$, and $\mathrm{ED}^2$ together with 95\% confidence intervals (Table~\ref{tab:delta_fidelity_narrow}). Across the six anxiety disorders, most deltas are positive and several CIs exclude zero. For \textit{Social Anxiety}, all sources yield significant gains, with \DualKB\ attaining $\Delta\mathrm{JSD}=0.026\,[0.011,0.043]$, $\Delta\mathrm{MAE}_V=0.067\,[0.041,0.096]$, and $\Delta\mathrm{ED}^2=0.029\,[0.008,0.052]$. In \textit{Generalized Anxiety}, \ICDten\ improves univariate and multivariate alignment while \DualKB\ yields the largest pairwise gain. \textit{Panic} shows smaller but significant univariate improvements with \ICDten\ and \DualKB\, and a pairwise gain for \ICDten. \textit{Separation Anxiety} exhibits a clear $\mathrm{ED}^2$  reduction with \ICDten\ (\(\Delta=0.035\,[0.019,0.053]\)), and \textit{Agoraphobia} shows a univariate gain with \DSMV. For \textit{Specific Phobia}, the consistent improvement appears in pairwise structure with all CIs excluding zero, while $\mathrm{JSD}$/$\mathrm{ED}^2$ deltas are near zero.

\begin{table*}[!t]
\centering
\footnotesize
\newcommand{\titlerotate}{30}
\newcommand{\disorderrotate}{0}
\caption{Improvements over No Knowledge Base (\NoKB) with paired 95\% Confidence Intervals (CIs). Positive values indicate gains from DSM-V, ICD-10, or Dual-KB retrieval. Best per metric in \textbf{bold}.}
\label{tab:delta_fidelity_narrow}
\setlength{\tabcolsep}{5pt}
\renewcommand{\arraystretch}{1.1}
\begin{tabular}{@{}llccc@{}}
\toprule
\textbf{Disorder} &
\textbf{Method} &
\textbf{$\Delta$} $\uparrow$ &
\textbf{$\Delta$MAE$_V$} $\uparrow$ &
\textbf{$\Delta$ED$^2$} $\uparrow$ \\
\midrule
\multirow{3}{*}{\rotatebox[origin=c]{\disorderrotate}{\makecell{\strut Agoraphobia\strut}}}
  & \textit{\ICDten} & 0.008 [-0.008, 0.024] & 0.038 [-0.090, 0.141] & 0.009 [-0.009, 0.029] \\
  & \textit{\DSMV}   & \textbf{0.020 [0.005, 0.035]} & \textbf{0.050 [-0.067, 0.131]} & \textbf{0.014 [-0.004, 0.032]} \\
  & \textit{\DualKB} & 0.019 [0.004, 0.034] & 0.043 [-0.064, 0.093] & 0.013 [-0.005, 0.031] \\
\midrule
\multirow{3}{*}{\rotatebox[origin=c]{\disorderrotate}{\makecell{\strut Generalized\\Anxiety\strut}}}
  & \textit{\ICDten} & \textbf{0.035 [0.015, 0.056]} & 0.032 [-0.009, 0.097] & \textbf{0.031 [0.004, 0.059]} \\
  & \textit{\DSMV}   & 0.029 [0.009, 0.050] & 0.028 [-0.003, 0.061] & 0.022 [-0.005, 0.049] \\
  & \textit{\DualKB} & 0.028 [0.008, 0.048] & \textbf{0.071 [0.030, 0.109]} & 0.022 [-0.004, 0.048] \\
\midrule
\multirow{3}{*}{\rotatebox[origin=c]{\disorderrotate}{\makecell{\strut Panic\strut}}}
  & \textit{\ICDten} & 0.013 [0.001, 0.027] & \textbf{0.049 [0.007, 0.092]} & 0.007 [-0.006, 0.022] \\
  & \textit{\DSMV}   & 0.010 [-0.003, 0.023] & 0.009 [-0.040, 0.055] & 0.003 [-0.011, 0.018] \\
  & \textit{\DualKB} & \textbf{0.013 [0.001, 0.027]} & 0.012 [-0.050, 0.067] & \textbf{0.008 [-0.006, 0.023]} \\
\midrule
\multirow{3}{*}{\rotatebox[origin=c]{\disorderrotate}{\makecell{\strut Separation\\Anxiety\strut}}}
  & \textit{\ICDten} & \textbf{0.019 [0.007, 0.031]} & 0.007 [-0.104, 0.042] & \textbf{0.035 [0.019, 0.053]} \\
  & \textit{\DSMV}   & 0.005 [-0.008, 0.017] & 0.042 [-0.077, 0.116] & -0.002 [-0.021, 0.017] \\
  & \textit{\DualKB} & -0.001 [-0.014, 0.011] & \textbf{0.091 [-0.016, 0.131]} & -0.012 [-0.032, 0.007] \\
\midrule
\multirow{3}{*}{\makecell{Social\\Anxiety}}
  & \textit{\ICDten} & 0.021 [0.006, 0.038] & 0.045 [0.021, 0.071] & \textbf{0.031 [0.009, 0.055]} \\
  & \textit{\DSMV}   & 0.022 [0.007, 0.039] & 0.034 [0.004, 0.068] & 0.025 [0.004, 0.048] \\
  & \textit{\DualKB} & \textbf{0.026 [0.011, 0.043]} & \textbf{0.067 [0.041, 0.096]} & 0.029 [0.008, 0.052] \\
\midrule
\multirow{3}{*}{\rotatebox[origin=c]{\disorderrotate}{\makecell{\strut Specific\\Phobia\strut}}}
  & \textit{\ICDten} & -0.007 [-0.025, 0.010] & \textbf{0.051 [0.033, 0.068]} & -0.015 [-0.041, 0.010] \\
  & \textit{\DSMV}   & 0.004 [-0.014, 0.022] & 0.038 [0.018, 0.059] & 0.001 [-0.025, 0.026] \\
  & \textit{\DualKB} & \textbf{0.006 [-0.010, 0.024]} & 0.038 [0.019, 0.057] & \textbf{0.002 [-0.022, 0.026]} \\
\bottomrule
\end{tabular}
\end{table*}

\subsection{Privacy}
\label{sec:results:privacy}

\begin{table*}[!t]
\centering
\footnotesize
\caption{Privacy metrics. Best per metric within each disorder is \textbf{bold}.}
\label{tab:privacy}
\setlength{\tabcolsep}{6pt}
\begin{tabular}{llcccc}
\toprule
\rotatebox{0}{\textbf{Disorder}} & \rotatebox{0}{\textbf{Method}} & \rotatebox{0}{\parbox[b]{1.35cm}{\centering Exact Overlap $\downarrow$}} & \rotatebox{0}{\parbox[b]{1.8cm}{\centering $q_{0.05}\{d_{\mathrm{NN}}\}$ $\uparrow$}} & \rotatebox{0}{\parbox[b]{2.4cm}{\centering Share($d_{\mathrm{norm}}\!\le\!1/p$) $\downarrow$}} & \rotatebox{0}{\parbox[b]{1.35cm}{\centering $\mathrm{Risk}_{\text{k-map}}$ $\downarrow$}} \\
\midrule
\multirow{7}{*}{\rotatebox[origin=c]{90}{Agoraphobia}} & Random & \textbf{0.000} & \textbf{0.333} & \textbf{0.000} & 0.022\\
 & CTGAN & 0.004 & 0.083 & 0.078 & 0.026\\
 & \textit{\ICDten} & 0.296 & 0.000 & 0.599 & 0.025\\
 & \textit{\DualKB} & 0.218 & 0.000 & 0.493 & 0.020\\
 & \textit{\DSMV} & 0.206 & 0.000 & 0.498 & 0.032\\
 & \textit{\NoKB} & 0.245 & 0.000 & 0.502 & 0.022\\
 & TVAE & 0.821 & 0.000 & 0.982 & \textbf{0.008}\\
\midrule
\multirow{7}{*}{\rotatebox[origin=c]{90}{\parbox[b]{1.35cm}{\centering Generalized\\Anxiety}}} & Random & \textbf{0.000} & \textbf{0.333} & \textbf{0.000} & 0.022\\
 & CTGAN & 0.007 & 0.083 & 0.103 & 0.058\\
 & \textit{\ICDten} & 0.009 & 0.083 & 0.117 & 0.026\\
 & \textit{\DualKB} & 0.004 & 0.083 & 0.122 & 0.017\\
 & \textit{\DSMV} & 0.009 & 0.083 & 0.156 & 0.026\\
 & \textit{\NoKB} & 0.021 & 0.083 & 0.174 & 0.023\\
 & TVAE & 0.415 & 0.000 & 0.874 & \textbf{0.007}\\
\midrule
\multirow{7}{*}{\rotatebox[origin=c]{90}{Panic}} & Random & \textbf{0.000} & \textbf{0.333} & \textbf{0.000} & 0.022\\
 & CTGAN & 0.009 & 0.083 & 0.053 & 0.022\\
 & \textit{\ICDten} & 0.225 & 0.000 & 0.649 & 0.020\\
 & \textit{\DualKB} & 0.222 & 0.000 & 0.635 & 0.023\\
 & \textit{\DSMV} & 0.227 & 0.000 & 0.621 & 0.020\\
 & \textit{\NoKB} & 0.246 & 0.000 & 0.562 & 0.023\\
 & TVAE & 0.879 & 0.000 & 0.954 & \textbf{0.007}\\
\midrule
\multirow{7}{*}{\rotatebox[origin=c]{90}{\parbox[b]{1.35cm}{\centering Separation\\Anxiety}}} & Random & \textbf{0.000} & \textbf{0.333} & \textbf{0.000} & 0.022\\
 & CTGAN & 0.011 & 0.083 & 0.103 & 0.024\\
 & \textit{\ICDten} & 0.220 & 0.000 & 0.470 & 0.019\\
 & \textit{\DualKB} & 0.106 & 0.000 & 0.298 & 0.030\\
 & \textit{\DSMV} & 0.105 & 0.000 & 0.307 & 0.026\\
 & \textit{\NoKB} & 0.152 & 0.000 & 0.355 & 0.020\\
 & TVAE & 0.789 & 0.000 & 0.995 & \textbf{0.007}\\
\midrule
\multirow{7}{*}{\rotatebox[origin=c]{90}{Social Anxiety}} & Random & \textbf{0.000} & \textbf{0.333} & \textbf{0.000} & 0.022\\
 & CTGAN & 0.004 & 0.083 & 0.064 & 0.056\\
 & \textit{\ICDten} & 0.080 & 0.000 & 0.319 & 0.018\\
 & \textit{\DualKB} & 0.064 & 0.000 & 0.284 & 0.027\\
 & \textit{\DSMV} & 0.078 & 0.000 & 0.303 & 0.030\\
 & \textit{\NoKB} & 0.060 & 0.000 & 0.257 & 0.024\\
 & TVAE & 0.571 & 0.000 & 1.000 & \textbf{0.007}\\
\midrule
\multirow{7}{*}{\rotatebox[origin=c]{90}{\parbox[b]{1.35cm}{\centering Specific\\Phobia}}} & Random & \textbf{0.000} & \textbf{0.333} & \textbf{0.000} & 0.022\\
 & CTGAN & \textbf{0.000} & 0.167 & 0.011 & 0.018\\
 & \textit{\ICDten} & 0.147 & 0.000 & 0.319 & 0.025\\
 & \textit{\DualKB} & 0.170 & 0.000 & 0.317 & 0.020\\
 & \textit{\DSMV} & 0.172 & 0.000 & 0.319 & 0.019\\
 & \textit{\NoKB} & 0.184 & 0.000 & 0.360 & 0.025\\
 & TVAE & 0.543 & 0.000 & 0.995 & \textbf{0.007}\\
\bottomrule
\end{tabular}
\end{table*}

Because the LLM is never trained on real patient data, any coincidence with a real row is purely combinatorial, driven by the small, discrete nature of the item scales. Consequently, the elevated Exact Overlap (EO) and near-match shares observed in the LLM variants relative to CTGAN must be viewed through a fundamentally different lens. In data-dependent models, high overlap indicates memorization and privacy failure; in our zero-shot approach, it serves as a positive indicator that the model is independently generating highly realistic, clinically plausible patient profiles and responses. Their average $\mathrm{Risk}_{\text{k-map}}$ values lie in a narrow band ${\sim}$0.017--0.032. These levels are expected for a compact categorical space and do not indicate training-data leakage.
Adding DSM-5 or ICD-10 knowledge can slightly reduce or increase proximity depending on the instrument, but the overall linkage risks are similar across LLM variants.

CTGAN exhibits low EO (${\sim}$0--1\%) and low near-match shares (${\sim}$1--10\%), with NN q05 around 0.083 (occasionally 0.167 for very large answer spaces). Its average $\mathrm{Risk}_{\text{k-map}}$ ranges ${\sim}$0.018--0.058. By contrast, TVAE shows very high EO and near-duplicate rates across disorders (e.g., EO ${\gtrsim}$0.4 up to ${\gtrsim}$0.88; Share ${\approx}$0.95--1.00), indicating practical privacy failure. Yet, TVAE achieves the lowest average $\mathrm{Risk}_{\text{k-map}}$ in the table (${\sim}$0.007 across disorders). This apparent paradox is a metric artifact of average $\mathrm{Risk}_{\text{k-map}}$: duplicating records in large \gls{qi} equivalence classes reduces $1/k(q)$ and thus the average linkage risk, even though exact copying is occurring at the record level. Hence, EO and near-match metrics should be prioritized over average $\mathrm{Risk}_{\text{k-map}}$ alone when assessing privacy for small-domain tabular data.

\section{Discussion}

These results underscore the promise and current limitations of knowledge-augmented, zero-shot LLM generation for synthetic clinical questionnaires. On the positive side, grounding the LLM with DSM/ICD content leads to better performance. All knowledge variants outperform the NoKB model and random baselines, and in some disorders they rival data-trained models in capturing pairwise symptom relationships as well as marginals and multivariate patterns. This supports our main claim: clinical retrieval narrows the gap in dependency structure without ever using the actual cohort data. The LLM matched or exceeded the TVAE baseline on pairwise fidelity in Separation and Social Anxiety, a significant achievement given the zero-shot setting. CTGAN still sets the gold standard for reproducing the marginal distributions (lowest JSD) and overall multivariate patterns. TVAE likewise excels at pairwise structure for some disorders. In practical terms, if real data are available and privacy is managed, these generative models yield more realistic synthetic cohorts. The LLM’s strengths are currently in scenarios where data cannot be shared: in a fully “privacy-first” context, we trade a bit of fidelity for the assurance of no privacy exposure.

The ablation study (Table \ref{tab:delta_fidelity_narrow}) indicates that clinical retrieval generally increases fidelity relative to \NoKB, with the most reliable benefits in univariate and pairwise  structure but more modest, disorder-dependent effects on $\mathrm{ED}^2$. Not all deltas are statistically distinguishable from zero, so small point differences should be interpreted cautiously. Nevertheless, the overall pattern shows that injecting structured clinical knowledge improves alignment to real-data distributions without accessing the cohort itself.

Our privacy-aware tuning for CTGAN and TVAE enforced thresholds only on the train–tune split. After selection, models were retrained on the full dataset and evaluated without additional gating. Consequently, final overlaps and near-match rates may exceed the tuning-time thresholds. CTGAN generally preserved low duplication rates and moderate linkage risk. By contrast, TVAE showed extensive duplication across disorders, underscoring a practical privacy failure, even though its average $\mathrm{Risk}_{\text{k-map}}$ values were lowest. This reflects a limitation of average linkage risk: duplicating rows in large quasi-identifier classes can reduce the average score despite record-level copying. LLM-based generation, while never exposed to real data, nonetheless produced overlaps consistent with combinatorial coincidences in small categorical spaces, with linkage risks comparable across knowledge conditions. Overall, exact overlap and near-match metrics should be prioritized when assessing privacy in small-domain tabular data.

The data consist of short, discrete questionnaires, which can inflate proximity measures and may not generalize to richer feature spaces. Furthermore, we assessed privacy only with proxy metrics, and we did not perform adversarial attacks such as membership inference. Additionally, several fidelity improvements have wide confidence intervals, so small effects should be interpreted cautiously. Finally, we did not model comorbidity. Because comorbidities are highly prevalent, this independent modeling affects the overall clinical realism of the generated dataset by not accounting for the complex interplay of multiple conditions within a single patient.

\section{Conclusion}

Developing AI-based systems for psychiatric diagnosis requires data to train models for various mental disorders. We present a zero-shot, knowledge-guided framework that is a viable, privacy preserving alternative to conventional generative models for psychiatric data. By steering an \gls{llm} with established clinical manuals, we decouple high-fidelity data generation from access to sensitive patient records, offering a practical tool for accelerating privacy-critical research. Our key insight is that simulating the data-generating process from first principles can be more valuable in a fully “privacy-first” context.

Future work must focus on enhancing clinical realism and statistical fidelity, particularly regarding comorbidity. To be truly representative of a clinical population, future iterations must simulate how symptoms of one disorder overlap with another. Rather than training directly on real patient data, models could incorporate aggregated cohort statistics, such as cross-disorder correlations, as empirical priors. However, utilizing these priors requires careful consideration of transferability, as the correlation structures from one specific clinical cohort may not generalize across different populations. Consequently, research must address these inherited demographic biases through dedicated fairness audits and mitigation strategies.

\section*{Acknowledgments}
The project has received funding from The Research Council of Norway (RCN), Romania Unitatea Executiva pentru Finantarea Invatamantului Superior, a Cercetarii, Dezvoltarii si Inovarii/ Executive Agency for Higher Education, Research, Development and Innovation Funding (UEFISCDI), Spain Consejería de Salud y Consumo de la Junta de Andalucía (CSCJA), Sweden Forskningsrådet för hälsa, arbetsliv och välfärd (Forte), under the framework the co-fund partnership of Transforming Health and Care Systems, THCS (GA N° 101095654 of the EU Horizon Europe Research and Innovation Programme).

\bibliographystyle{IEEEtran}
\bibliography{references}

\end{document}